\documentclass[sigconf]{aamas} 
\usepackage{balance, multirow, amsmath} 

\title{Automating Credit Card Limit Adjustments \texorpdfstring{\\}{} Using Machine Learning}

\author{Diego Pestana}
\affiliation{
  \institution{Venezolano de Crédito}
  \city{Caracas}
  \country{Venezuela}}
\email{dpestana@venezolano.com}

\author{Enrique Areyan}
\affiliation{
  \institution{enriqueareyan.com}
  \city{Seattle}
  \country{USA}}
\email{enrique3@gmail.com}

\begin{abstract}
Venezuelan banks have historically made credit card limit adjustment decisions manually through committees. However, since the number of credit card holders in Venezuela is expected to increase in the upcoming months due to economic improvements, manual decisions are starting to become unfeasible. In this project, a machine learning model that uses cost-sensitive learning is proposed to automate the task of handing out credit card limit increases. To accomplish this, several neural network and XGBoost models are trained and compared, leveraging Venezolano de Credito’s data and using grid search with 10-fold cross-validation. The proposed model is ultimately chosen due to its superior balance of accuracy, cost-effectiveness, and interpretability. The model’s performance is evaluated against the committee’s decisions using Cohen’s kappa coefficient, showing an almost perfect agreement.
\end{abstract}

    
\newcommand{\BibTeX}{\rm B\kern-.05em{\sc i\kern-.025em b}\kern-.08em\TeX}

\setcopyright{none} 
\renewcommand\footnotetextcopyrightpermission[1]{}
\begin{document}

\pagestyle{fancy}
\fancyhead{}

\settopmatter{printacmref=false}

\maketitle 
\begin{sloppypar}

\section{Introduction}
In the last 10 years, financial instability has cut Venezuela's number of credit card holders ~\cite{Arm23}. Hyperinflation consumed the balances of previously emitted credit cards, and the high legal reserve requirements prevented banks from handing out new ones. The combination of these two resulted in a decrease in credit card holders and, consequently, credit card limit adjustment decisions (CLAD) made yearly. Venezuelan banks, such as Venezolano de Crédito (VDC), have traditionally made limit adjustment decisions manually through committees. However, Venezuela's recent economic improvement has increased the number of credit cards, making manual adjustments no longer possible. Thus, banks need a solution to automate limit adjustments using minimal human intervention, preventing dissatisfaction from long waits and biased decisions.

This project proposes a way to automate credit card limit adjustments, aiming to avoid manual and biased evaluations, and taking into consideration misclassification costs. By leveraging automation, more frequent and data-driven CLAD can be made, resulting in faster wait times for clients and a potential revenue increase. Multiple statistical methods have been used, including machine learning (ML), to adjust credit card limits. ~\cite{Lic21} used a Markov decision process to determine the optimal dynamic credit limit policy. Several studies ~\cite{Gui19, Mun2019} evaluated and compared multiple ML models\footnote{Including logistic regression, random forest, XGBoost, and neural networks} to predict customers' behavior in paying off credit card balances, suggesting that neural networks (NN) and XGBoost are the most effective type of models.

First, I will train and evaluate a range of NN and XGBoost models using real-world data\footnote{VDC’s CLAD made over the last five years} and cost-sensitive learning to determine whether a credit card holder should receive a limit adjustment. Based on the results, I will select a model that automatically adjusts credit card limits while requiring little human intervention. Finally, to evaluate the proposed model’s performance, I will compare recent CLAD made by the committee against the model’s predictions. To quantify the level of agreement between the two, I will measure their inter-rater reliability using Cohen’s kappa coefficient. 

\section{Data Analysis}
A CLAD involves evaluating a client’s profile and historical data as input and determining whether to give a limit adjustment (positive outcome) or deny it (negative outcome).

I collected a dataset of 10,000 CLAD made over the last five years, from September 2019 to September 2024. Each CLAD has 13 attributes. To avoid bias and ensure fairness, demographic attributes like age and gender were intentionally excluded from the dataset. Thus, the dataset only contains objective and performance-based attributes, such as current limit balance, account longevity, and an agency's credit score. Due to privacy concerns and VDC restrictions, not all attributes are shown in this extended abstract.

A simple statistical analysis of the data set shows that the average limit balance is 1,463.72 bolívares (BS)\footnote{Approximately \$30 at the time of writing}, the median credit rating is BB\footnote{The agency’s credit system goes by AAA, AA, A, BBB, BB, B, CCC, CC, C, and D; with BB considered "poor"}, the youngest account is two years old and the oldest one is 22 years old. Additionally, 7,454 instances are positive, while 2,546 are negative.

\section{Models}
Venezuelan banks hand out credit card limit adjustments nowadays through committees. For example, VDC's risk management committee has a bi-weekly meeting to make CLAD for a randomly selected group of card holders. Before evaluating each one, the committee sets a fixed adjustment rate \(\alpha\) to be applied to all positive outcomes.

The models' goal is to determine whether an ~\(\alpha\) percentage credit card limit adjustment is given, making it a supervised learning binary classification task. They are meant to be run offline before or during each committee meeting. I will train and compare two types of models: NN and XGBoost.

A NN model has three parts: (1) an input layer, (2) one or more hidden layers, and (3) an output layer that returns the prediction. The NN propagates data through linear transformations between layers, using non-linear activation functions within each layer to compute the output ~\cite{Alp10}. It is trained to minimize a loss function, typically the root mean squared error ~\cite{Gol88}. Among the different ML models used for classification, NN are considered state-of-the-art, typically achieving better performance ~\cite{Alp10}.

An XGBoost model is an advanced implementation of the gradient boosting decision trees algorithm (GBDT). GBDT builds a model from multiple weak decision trees, combining them using gradient boosting to create a stronger one. These types of models are trained similarly to NN, placing more weight on instances with erroneous predictions to gradually minimize a loss function ~\cite{Che16}.

A typical classification model assumes that misclassification costs (false negative and false positive costs) are the same ~\cite{Tha10}. However, for the task of handing out credit card limit adjustments, this assumption is not true because denying an adjustment to a good client is more tolerable than giving one to a bad client who may never repay it. For this reason, I decided to use a cost-sensitive learning approach. 

In cost-sensitive learning, instead of each instance being either correctly or incorrectly classified, each class (false positive or false negative) is given a misclassification cost. Thus, instead of optimizing accuracy, the goal is to minimize the total misclassification cost ~\cite{HeMa13}.

In this case, the misclassification cost of a false positive \(C_{FP}\) is equivalent to the client defaulting his entire credit card limit. On the other hand, the misclassification cost of a false negative \(C_{FN}\) is the loss in profit, which is estimated to be the minimum payment of the first month plus an additional administrative cost. In contrast, correct classification costs are assumed to be nonexistent.

\begin{figure}[h!]
  \centering
  \includegraphics[width=\linewidth]{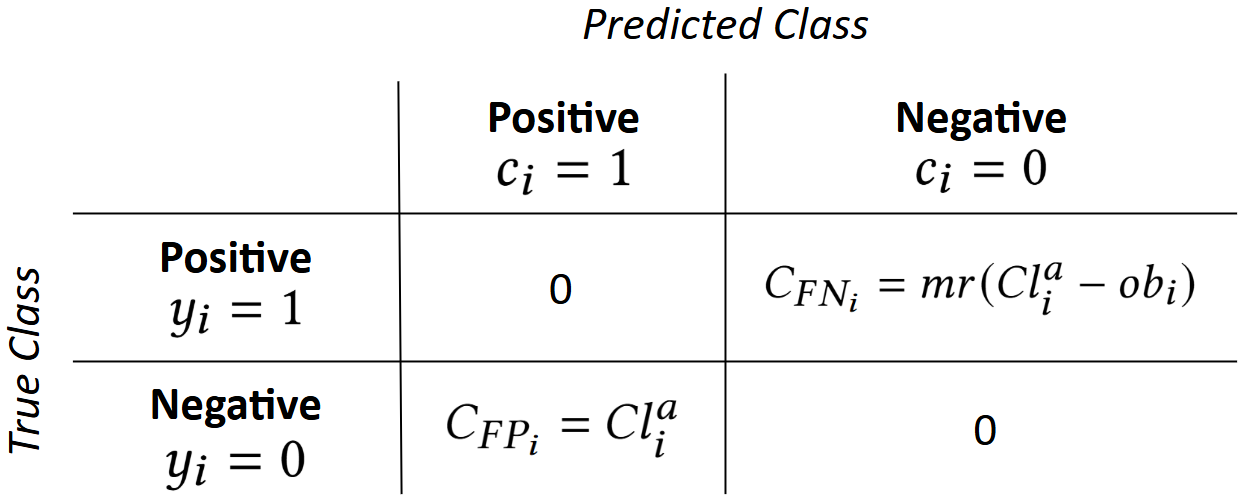}
  \caption{Classification cost matrix}
  \label{fig:cost-matrix}
  \Description{Classification cost matrix}
\end{figure}

Moreover, in this case the costs vary not only among classes but also instances, since credit limits vary between clients. Methods where the misclassification cost varies between instances are called instance-dependent methods. Leveraging an instance-dependent cost matrix for credit scoring ~\cite{Cor15}, the cost matrix shown in Figure ~\ref{fig:cost-matrix} is built for this task, where \(mr\) is the minimum payment percentage, \(Cl_i^a\) is the credit card limit after adjustment of client \(i\), and \( ob_i\) is the outstanding balance of client \(i\).

The credit card limit after adjustment is determined by \eqref{eq:claa}, where \(Cl_i^b\) is the credit card limit before adjustment of client \(i\). The cost function \(C\) defined for all models is \eqref{eq:cf}.

\begin{eqnarray}\label{eq:claa}
Cl_i^a = Cl_i^b(1 + \alpha)
\end{eqnarray}

\vspace{-0.75\baselineskip}
\begin{eqnarray}\label{eq:cf}
C & = & \sum_{i = 1}^{m} y_i(1-c_i)C_{FN_i} + c_i(1-y_i)C_{FP_i}
\end{eqnarray}

NN models have many parameters and hyperparameters. These include the number of hidden layers, neurons per layer, activation function, gamma, and learning rate. The XGBoost model has parameters and hyperparameters too, like maximum depth, minimum child weight, maximum number of leaves, and learning rate. Manually tuning each hyperparameter is time-consuming, so I applied grid search with k-fold cross-validation.

Grid search is an algorithm used for hyperparameter tuning. It exhaustively tries every combination of a provided hyperparameter search space to find the best model ~\cite{Lia19}. K-fold cross-validation is a resampling procedure that splits the dataset into k bins, and uses one of them for testing and the rest for training. It is very useful for small datasets—like the one at hand—as it uses as much data as possible both for testing and training ~\cite{Jam13}. The process is repeated k times, alternating the testing bin.

I used k = 10, as it has been shown empirically to suffer neither from excessively high bias nor very high variance ~\cite{Jam13}. The NN and XGBoost models’ search spaces used for the grid search with 10-fold cross-validation can be found in Appendix ~\ref{sec:appendix-a}.

I calculated the cost of every possible combination of hyperparameters. Ultimately, I chose the hyperparameters that resulted in the lowest cost, for both NN and XGBoost. 

\section{Results and Discussion}
\begin{figure}[h!]
  \centering
  \includegraphics[width=0.75\linewidth]{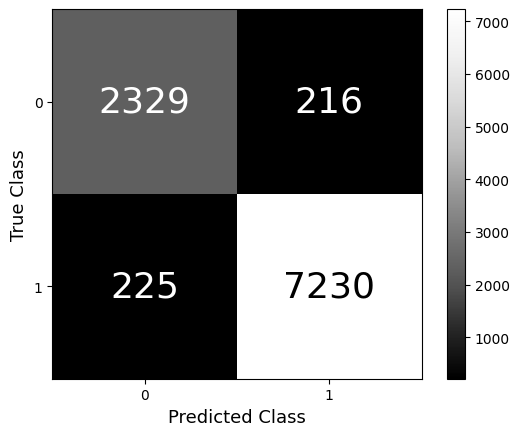}
  \caption{NN model’s confusion matrix}
  \label{fig:nn-conf-matrix}
  \Description{NN model’s confusion matrix}
\end{figure}

The top-performing NN model has four hidden layers (4, 4, 6, and 8 neurons) using ReLU and L2 regularization. The trained NN model predicts 95.59\% of test cases correctly, and returns a total cost of 834,988.96 BS\footnote{Roughly \$16,058 at the time of writing}. Figure ~\ref{fig:nn-conf-matrix} shows that there are 7,230 true positives and 2,329 true negatives.

\begin{figure}[ht!]
  \centering
  \includegraphics[width=0.75\linewidth]{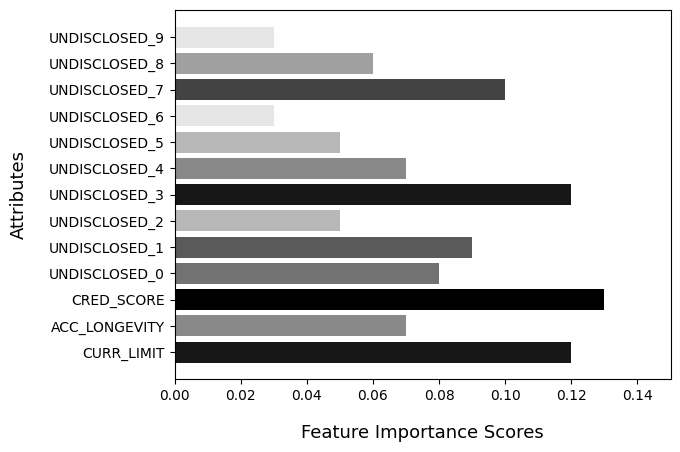}
  \caption{Feature importance scores}
  \label{fig:xg-imp}
  \Description{Feature importance scores}
\end{figure}

The XGBoost model with the best results is the one with 10 trees, a maximum depth of 6, and a minimum child weight of 3. Figure ~\ref{fig:xg-imp}  helps visualize the features that play a significant role in predicting the output.

\begin{figure}[h!]
  \centering
  \includegraphics[width=0.75\linewidth]{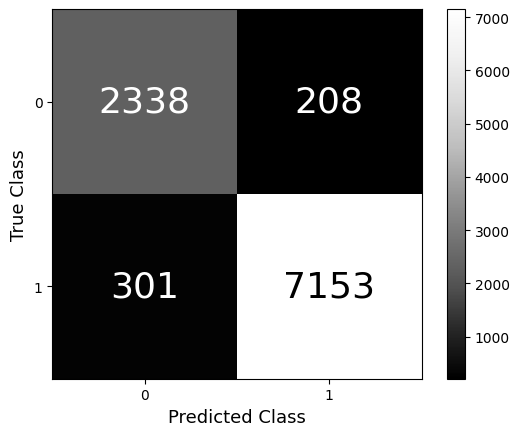}
  \caption{XGBoost model’s confusion matrix}
  \label{fig:xg-conf-matrix}
  \Description{XGBoost model’s confusion matrix}
\end{figure}

The trained XGBoost model predicts 94.91\% of test cases correctly, and returns a total cost of 809,660.81 BS\footnote{Around \$15,570 at the time of writing}. Figure ~\ref{fig:xg-conf-matrix} shows that there are 7,153 true positives and 2,338 true negatives.

Even though the NN model was 0.68\% more accurate than the XGBoost model, the number of false positive labels predicted—clients defaulting their entire credit card limit—was lower for the XGBoost model, 208 to 216. Thus, the XGBoost model is 25,328.15 BS cheaper than the NN model, since false negatives—loss in profit—do not have as big an impact on cost as false positives.

Additionally, the complex nature of the NN model made it difficult to understand, essentially being a black box. In banking, regulators require clear explanations for all decisions, making the NN model unfeasible. In contrast, tree-based ML models are used in domains such as finance, where interpretability is important ~\cite{Lun19}, as they consistently outperform standard NN models on tabular-style datasets where features are individually meaningful and do not have strong multi-scale temporal structures ~\cite{Che16}. Since the XGBoost model offers a better balance of accuracy and interpretability, it is the proposed solution. Leveraging ~\cite{Lun19} Tree Explainer algorithm, the decisions made by the XGBoost model could be explained to regulators using game theory.  

To analyze the proposed model’s performance, I took the CLAD made during the first half of October by the committee and compared them to the model’s predictions. Through Cohen’s kappa coefficient ~\(\kappa\), an inter-rater reliability measure, I calculated the degree of agreement among the committee and the model. I used ~\(\kappa\) since I only have two raters—the committee and the model—and because it takes into account the possibility of agreement occurring by chance, making it a more robust measure than simple percent agreement calculation. The formula for calculating k is:

\begin{eqnarray}\label{eq:kappa}
\kappa = \dfrac{P_0 + P_e}{e}
\end{eqnarray}

The proportion of times the committee and the model agreed is ~\(P_0\), and it is calculated using \eqref{eq:p0} where  ~\(TP\) is the number of true positives—the adjustment was given by the two of them—,  ~\(TN\) is the number of true negatives—the adjustment was not given by any of them—, and  ~\(N\) is the total number of CLAD.

\begin{eqnarray}\label{eq:p0}
P_0 = \dfrac{TP - TN}{N}
\end{eqnarray}

In contrast, ~\(P_e\) is the probability that both the committee and the model would make the same decision if they guessed randomly. Thus, it is the sum of the probability that they randomly give the limit adjustment \eqref{eq:p1} and the probability that they do not \eqref{eq:p2}.

\begin{eqnarray}\label{eq:p1}
P_1 = \dfrac{(TP + FN) \cdot (TP + FP)}{N^2}
\end{eqnarray}

\begin{eqnarray}\label{eq:p2}
P_2 = \dfrac{(TN + FN) \cdot (TN + FP)}{N^2}
\end{eqnarray}

To determine the number of false negatives ~\((FN)\) and false positives ~\((FP)\), since it is not clear which rater is objectively correct, I arbitrarily define ~\(FP\) as the number of CLAD the committee gave but the model did not, and ~\(FN\) as the number of CLAD the model gave but the committee did not.

\begin{figure}[h!]
  \centering
  \includegraphics[width=0.85\linewidth]{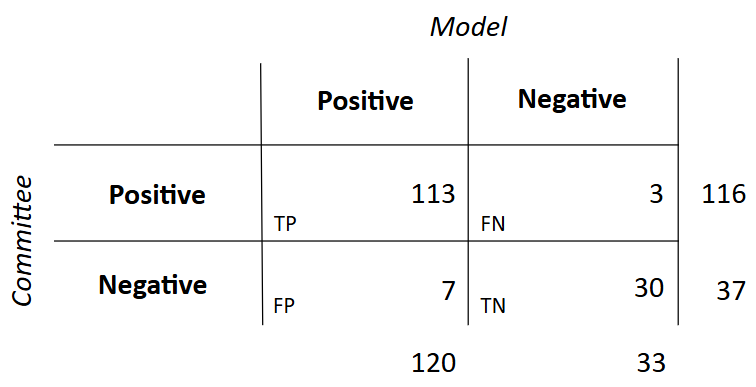}
  \caption{First half of October CLAD’s confusion matrix}
  \label{fig:oct-conf-matrix}
  \Description{First half of October CLAD’s confusion matrix}
\end{figure}

During the first half of October, the committee made 153 CLAD: 116 were positive and 37 were negative. After running the model over the same 153 instances, it predicted 120 positive outcomes and 33 negative outcomes. Figure ~\ref{fig:oct-conf-matrix} illustrates the comparison between the committee’s and model’s assessments, showing the frequency they agreed/disagreed on.

Using ~\eqref{eq:kappa}, I calculated Cohen’s kappa coefficient, getting ~\(\kappa = 0.81\). Cohen’s kappa coefficient ranges anywhere from 1 (all CLAD are the same) to -1 (all CLAD are different) ~\cite{Car96}. According to Cohen’s interpretation, a ~\(\kappa\) ranging from 0.8 to 0.9 is considered an almost perfect agreement ~\cite{Ueb87}. 

Despite getting a ~\(\kappa\) value considered almost perfect, I meticulously analyzed those CLAD where the committee and the model disagreed. A particular ~\(FP\) caught my attention, where a committee member disclosed that the primary reason for denying the limit increase was the client’s age, claiming that younger clients are perceived as higher risk. However, since the model does not consider age or any demographic data to estimate its output, it considered the client eligible for a limit increase. On the other hand, each ~\(FN\) displayed a poor credit score and low credit limit, which is consistent with the model’s importance scores. Further investigation needs to be done, in collaboration with the committee, to better understand the cause of these discrepancies. 

In summary, the proposed XGBoost model has proven to be an effective solution for determining ~\(\alpha\) percentage credit card limit adjustments, achieving an accuracy comparable to the state-of-the-art NN model trained (with a difference of less than 1\%). Additionally, the XGBoost model offers greater interpretability, lower costs, and almost perfect agreement with the committee’s decisions.

This project's findings have enabled VDC to automate their CLAD, while maintaining transparency and explainability. Currently, VDC’s risk management committee is using the proposed model in their analysis. As the model continues to improve, it has the potential to positively impact other Venezuelan banks, providing a framework for automating their own CLAD. 

One of the main limitations of this project was the small amount of data available. Nonetheless, a significant increase in CLAD is expected, which will provide a much larger dataset. Likewise, the small dataset allowed for grid search with 10-fold cross-validation, but as the dataset increases in size, alternative methods for evaluating hyperparameters would need to be explored. Lastly, although the model is designed to operate offline, as it continues to improve over time, it could be run online if necessary. 

\balance{}

\bibliographystyle{ACM-Reference-Format} 
\bibliography{acla}

\appendix
\appendixpage
\section{Hyperparameter Search Spaces}
\label{sec:appendix-a}

In this appendix, I provide an overview of the hyperparameter search spaces used during the grid search with 10-fold cross-validation. Specifically, Table ~\ref{tab:search-space-nn} shows the hyperparameter search space for the NN models, while Table ~\ref{tab:search-space-xg} presents the one for the XGBoost models.

\begin{table}[h!]
  \caption{Search space for NN models}
  \label{tab:search-space-nn}
  \begin{tabular}{c c}
    \toprule
    \textit{Parameters} & \textit{Search Space} \\ 
    \midrule
    Batch size & 2, 4, 6, 8 \\
    Epochs & 1, 2, 3, 4, 5, 6, 7, 8, 9 \\
    Optimizer & Adam, SGD, RMSprop \\
    Activation & ReLU, sigmoid, tanh \\
    Unit & 0.1 \\
    Learning rate & 0.1 \\ 
    L2 & true, false \\ 
    Maximum delta step & 0.4, 0.6, 0.8, 1 \\
    Subsample ratio & 0.9, 0.95, 1 \\ 
    Column subsample ratio & 0.9, 0.95, 1 \\ 
    Gamma & 0, 0.001 \\ 
    \bottomrule
  \end{tabular}
\end{table}

\begin{table}[ht]
  \caption{Search space for XGBoost models}
  \label{tab:search-space-xg}
  \begin{tabular}{c c}
    \toprule
    \textit{Parameters} & \textit{Search Space} \\ 
    \midrule
    Maximum tree depth & 2, 3, 4, 5, 6, 7, 8, 9 \\
    Minimum child weight & 1, 2, 3, 4 \\
    Early stop round & 100 \\
    Maximum epoch number & 500 \\
    Learning rate & 0.1 \\
    Number of boost & 60 \\ 
    Maximum delta step & 0.4, 0.6, 0.8, 1 \\ 
    Subsample ratio & 0.9, 0.95, 1 \\
    Column subsample ratio & 0.9, 0.95, 1 \\
    Gamma & 0, 0.001 \\ 
    \bottomrule
  \end{tabular}
\end{table}

\end{sloppypar}
\end{document}